\documentclass[a4paper,11pt]{article} 

\usepackage[left=2.1cm,right=2.1cm,top=2.5cm,bottom=2.5cm]{geometry}
\usepackage{amsmath,amsfonts,amssymb}
\usepackage{graphicx}
\usepackage{xcolor} 
\usepackage{eurosym}

\usepackage{booktabs}
\usepackage{array}
\usepackage{multirow}
\usepackage{subcaption}
\usepackage{hyperref}
\hypersetup{
	pdftitle={Comparing Prophet and Deep Learning to ARIMA in Forecasting Wholesale Food Prices},
	pdfauthor={Lorenzo Menculini, Andrea Marini, Massimiliano Proietti, Alberto Garinei, Alessio Bozza, Cecilia Moretti and Marcello Marconi},
}

\usepackage[sorting=none, style=ieee]{biblatex}
\bibliography{forecasting_biblio}

\usepackage{caption}
\usepackage{authblk}

\title{Comparing Prophet and Deep Learning to ARIMA in Forecasting Wholesale Food Prices}

\author[1]{Lorenzo Menculini\footnote{Corresponding author: lmenculini@idea-re.eu}}
\author[1]{Andrea Marini}
\author[1]{Massimiliano Proietti}
\author[2,1]{Alberto Garinei}
\author[3]{Alessio~Bozza}
\author[]{Cecilia Moretti}
\author[2,1]{Marcello Marconi}
\affil[1]{Idea-re S.r.l., Perugia, Italy}
\affil[2]{Department of Engineering Sciences, Guglielmo Marconi University, Rome, Italy}
\affil[3]{Cancelloni Food Service S.p.A., Magione (PG), Italy}

\date{}

\begin{document}

\maketitle


\begin{abstract}
	\noindent  Setting sale prices correctly is of great importance for firms, and the study and forecast of prices time series is therefore a relevant topic not only from a data science perspective but also from an economic and applicative one. 
  In this paper we examine different techniques to forecast sale prices applied by an Italian food wholesaler, as a step towards the automation of pricing tasks usually taken care by human workforce.
  We consider ARIMA models and compare them to Prophet, a scalable forecasting tool by Facebook based on a generalized additive model, and to deep learning models exploiting Long Short--Term Memory (LSTM) and Convolutional Neural Networks (CNNs).
  ARIMA models are frequently used in econometric analyses, providing a good benchmark for the problem under study.
  Our results indicate that ARIMA models and LSTM neural networks perform similarly for the forecasting task under consideration, while the combination of CNNs and LSTMs attains the best overall accuracy, but requires more time to be tuned.
  On the contrary, Prophet is quick and easy to use, but considerably less accurate.
\end{abstract}

\section{Introduction}

The main aim of firms is profit maximization.
To achieve this goal, the constant updating and forecasting of selling prices is of fundamental importance for every company.
Although the digital transformation is a phenomenon that is involving all companies, from small to large, many of them still update prices by hand through logics that are not always clear nor objective and transparent, but rather based on the experience and expertise of those in charge of updating the price list.
On the other hand, the automation of price prediction and update can provide a strong productivity boost by freeing up human resources, which can thus be allocated to more creative and less repetitive tasks.
This also increases the morale and commitment of employees; it also speeds up the achievement of goals, and improves accuracy by minimizing human errors.
Subjectivity is also reduced: once the operating criteria have been established, forecast algorithms will keep behaving consistently.
This in turn means an improvement in compliance.

Besides the automation of price updates, the prediction of the sales prices charged to customers in the short term also holds great value.
In general, organizations across all sectors of industry must undertake business capacity planning to be efficient and competitive. 
Predicting the prices of products is tightly connected to demand forecasting and therefore allows for a better management of warehouse stocks.
The current economic crisis caused by COVID-19 has highlighted the value of such management optimization, stressing the importance of the companies’ ability to minimize inventory reserves and just-in-time production models.
Forecast models considered in this paper can contribute to keeping the difference between wholesale purchase prices and company's sales prices under control, in view of maximizing the gross operating income.
They can therefore help companies avoid the risk of incalculable losses and ultimately improve their contractual capacity.

The present work proposes to deal with these topics by investigating and comparing different price forecasting models.
The specific task we consider is that of predicting the prices of three food products sold by a medium/large-size local wholesaler based in central Italy.
In such way, we investigate the predictability of wholesale prices, comparing the performance of traditional econometrics time series forecasting models with Facebook's Prophet and machine learning models.
The main goal of this paper is therefore to develop a forecasting model that could represent a first step towards the automation of the price-setting process, thus effectively aiding the work of company employees.  
In this way, it aims to be of practical use to companies for the maintenance and management of price lists.
Scalability and flexibility of the models presented in this paper are also an important point: for the sake of simplicity, we have applied the models to three different products, but we underline that the same models and algorithms can be easily applied to any product.

Time series forecasting has always been a major topic in data science with plenty of applications.
For a general review of some of the most used tools, see for example~\cite{adhikari2013introductory}.
Well-known traditional econometric methods are not always appropriate to study and forecast big and noisy time series data. 
This has generated particular interest in machine learning methods, bolstering data driven approaches that include a wide range of methods that have the advantage of not relying on prior assumptions and knowledge on data. 
See for example~\cite{fawaz2019deep,sezer2020financial,lim2021time,lara2021experimental} for reviews focusing on the application of deep learning~\cite{lecun2015deep} to time series.
Long short--term memory (LSTM) networks~\cite{hochreiter1997long} and convolutional neural networks (CNNs)~\cite{lecun1995convolutional} are almost ubiquitous in time series forecasting with machine learning.
CNNs are even more commonly used for image recognition and feature extraction.
However, the forecasting accuracy of standalone CNNs can be relatively low~\cite{alibavsic2019new}. 

The literature concerning economic time series prediction employing various methos -- for classical to artificial intelligence ones -- is very rich.
Nevertheless, although we believe automatic updating mechanisms and forecasting of sale prices are of uttermost relevance, the literature on these topics is not as developed as one would expect.
Most studies focus primarily on the implementation of models for the analysis and forecasting of general price levels (inflation) or commodity and stock market prices.
The forecasting of food prices in China was considered by the authors of~\cite{haofei2007neural,zou2007investigation}. 
In particular, Zou et al.~\cite{zou2007investigation} compared the performances of ARIMA, neural networks (NNs) and a combination of the two to forecast wheat prices in the Chinese market.
Their findings showed that, overall, NNs perform best at the task.
Neural networks were also employed in~\cite{jha2013agricultural} to forecast monthly wholesale prices of two agricultural products.
Ahumada and Cornejo~\cite{ahumada2016forecasting} considered a similar problem, also taking into account possible cross-dependencies of different product prices. 
In~\cite{pavlyshenko2019machine} the author focused on sales forecasting using machine learning models, a topic similar to the one considered in the present paper.
For more recent work on forecasting commodities prices see~\cite{livieris2020cnn}, where the authors forecasted gold prices, and~\cite{zhang2018prediction} where the Levenberg-Marquardt Backpropagation (LM-BP) algorithm was applied to stock prices prediction.
Other authors used machine learning methods for inflation forecasting~\cite{THAKUR201687,paranhos2021predicting}, also in comparison with more classical econometric models~\cite{araujo2020machine}.
Xue et al.~\cite{yan2021research} recently presented a high-precision short-term forecasting model for financial market time series employing deep LSTM neural networks, comparing them with other NN models.
Their results showed that LSTM deep neural networks have high forecasting accuracy for stock market time series.
In 2020, Kamalov~\cite{kamalov2020forecasting} evaluated multilayer perceptrons, CNNs and LSTM neural networks to forecast significant changes in stock prices for four major US public companies, showing that these three methods yield better results when compared to similar studies that forecast the direction of price change.
For models similar to the ones considered in this work and applied again to stock indexes forecasting, see~\cite{hao2020predicting}.
Hybrid ARIMA/neural network models where instead studied by the authors of~\cite{xiao2012hybrid}.
Stock prices have also been forecasted using LSTMs in conjunction with the attention mechanism~\cite{qiu2020forecasting}.
Machine learning models using LSTMs and CNNs are of widespread use in time series forecasting, well beyond the financial and economic realm.
For recent work on time series forecasting using machine learning outside the economic and financial area see~\cite{chimmula2020time}, an application to COVID-19 spreading forecasting, and~\cite{wu2020deep} for an application of deep learning to Influenza prevalence forecasting.

In this paper we compare the performance of standard Autoregressive Integrated Moving Average (ARIMA) models~\cite{box2015time}, which we take as a benchmark, to Prophet -- a forecasting tool developed by Facebook and based on a Generative Additive Model (GAM)~\cite{hastie1987generalized} -- and machine learning models exploiting LSTMs, both on their own and in combination with CNNs.
ARIMA univariate models are considered a standard reference model in econometrics. 
The compared models are rather different, as are the datasets that they accept in input, making the comparison interesting.
On one hand, Prophet's driving principles are simplicity and scalability; it is specifically tailored for business forecasting problems and handles missing data very well by construction.
On the other, the NN models we construct allow for a multivariate regression, fully exploiting all the collected data, but also require some data pre-processing, as does ARIMA.
Prophet has been compared to ARIMA models for the prediction of stock prices~\cite{chan2020time} and bitcoin \cite{yenidougan2018bitcoin}.

Our results indicate that the combination of CNNs and LSTMs yields the most accurate results for all the three products, but require the longest and computationally more expensive tuning.
On the contrary, Prophet performances were not brilliant, but model tuning and data preparation were particularly quick.
ARIMA and LSTM-only neural networks showed good performances both in terms of accuracy and time required for model selection and training.

The rest of the paper proceeds as follows. 
Section~\ref{sec:methods} introduces the dataset features, discussing its properties and some pre-processing steps that were taken on it; it also briefly presents the three models under consideration, their set-up and tuning.
In Section~\ref{sec:results} we give the results obtained with the three approaches and compare them.
We conclude in Section~\ref{sec:discussion} by discussing the results of this study and providing an outlook on future perspectives in the light of the paper findings.

\section{Materials and Methods}
\label{sec:methods}

\subsection{Dataset description and preparation}
\label{ssec:dataprep}
For this study we had access to a dataset comprising a total of approximately 260,000 food order records, reporting the following information: \emph{date of order}, \emph{order number}, \emph{unit price}, \emph{article code}, \emph{sold quantity}, \emph{customer code}, \emph{offer} (if present) and \emph{offer type}, \emph{unitary cost}.  
The records were collected by the wholesaler in a period ranging from year 2013 to 2021.
For the study conducted in this paper, we decided to focus on the three products with the most records, namely \emph{Carnaroli rice 1kg $\times$ 10} (henceforth product 1), \emph{Gorgonzola cheese 1/8 of wheel 1.5 kg} (product 2) and \emph{Cured aged ham 6.5 kg} (product 3).
The forecasting task considered in this work was to predict the average selling price for the following week, for each of the selected products.

As a first thing, we chose to leave out all data following the outbreak of the COVID-19 pandemic.
This was motivated by the huge impact that the lockdowns and restrictions imposed by the authorities had on the food and catering sector, introducing a major shock in sales trends at all scales.
Therefore, we excluded all records dated later than March 9, 2020 (last day before the first national lockdown in Italy).

\begin{figure}[h]	
	\centering
\includegraphics[width=13 cm]{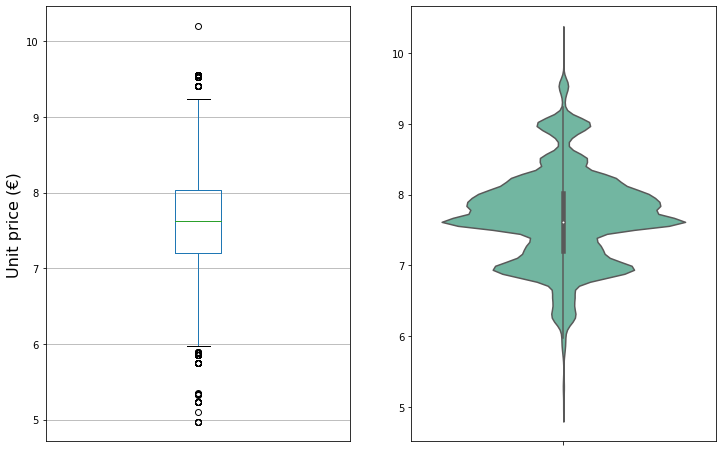}
\caption{Price distribution (after \emph{z-score} filtering) for product 2: to the left, boxplot, and to the right, violin plot.\label{fig:p1}}
\end{figure}

A preliminary data analysis revealed that the dataset contained a good number of outliers: for some of them, it appeared evident that this was due to incorrect typing of the product sale price.
To improve the quality of the dataset, we calculated the \emph{z-score} of each record based on its price as
$
z=(p-\bar{p}^{(w)})/\sigma^{(w)}\, ,
$
where $p$ is the unit sale price and $\bar{p}^{(w)}$ and $\sigma^{(w)}$ are the mean and standard deviation for the selected product, weighted by the quantity sold in each order.
Then, we filtered out all records with $|z|>4$.
Figure~\ref{fig:p1} shows the price distribution for product 2, after the filtering.

\begin{figure}[!h]
  \centering
  \begin{subfigure}[b]{0.48\textwidth}
    \centering
    \includegraphics[width=8.5 cm]{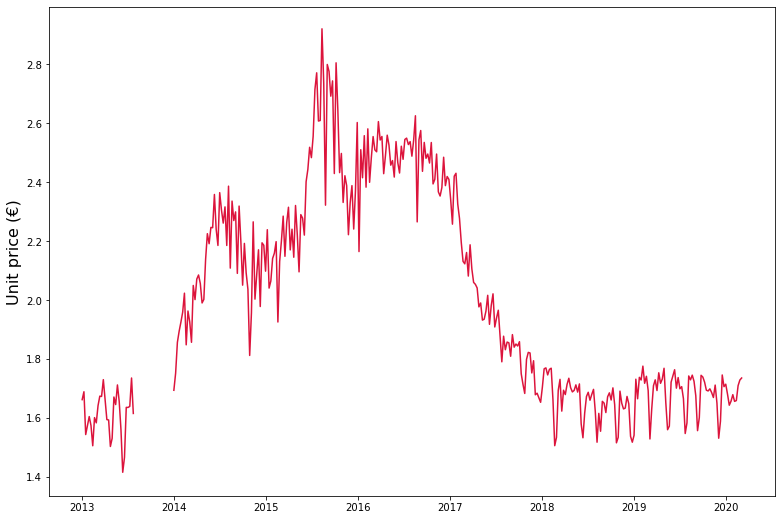}
  \end{subfigure}
  \hfill
  \begin{subfigure}[b]{0.48\textwidth}
    \centering
    \includegraphics[width=8.5 cm]{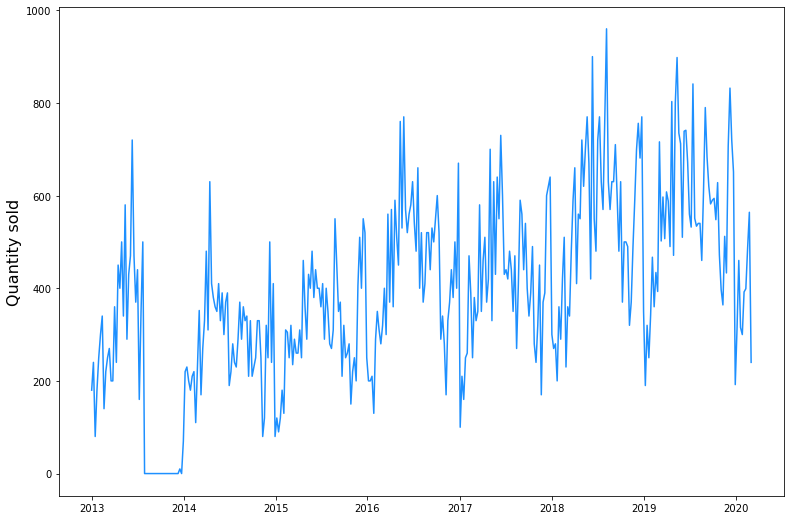}
  \end{subfigure}
  \caption{\textbf{(a)} Unit price and \textbf{(b)} sold quantity time series for product 1 after resampling with weekly frequency.\label{fig:p1_ts}}
\end{figure}

\begin{figure}[!h]
  \centering
  \begin{subfigure}[b]{0.48\textwidth}
    \centering
    \includegraphics[width=8.5 cm]{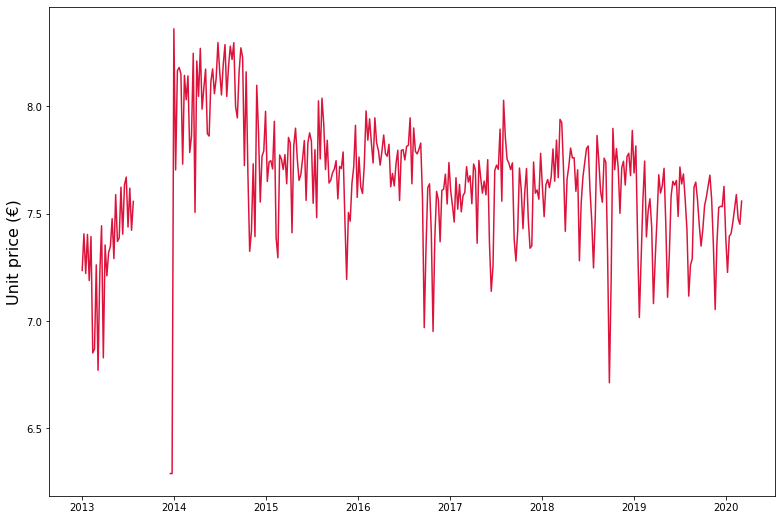}
  \end{subfigure}
  \hfill
  \begin{subfigure}[b]{0.48\textwidth}
    \centering
    \includegraphics[width=8.5 cm]{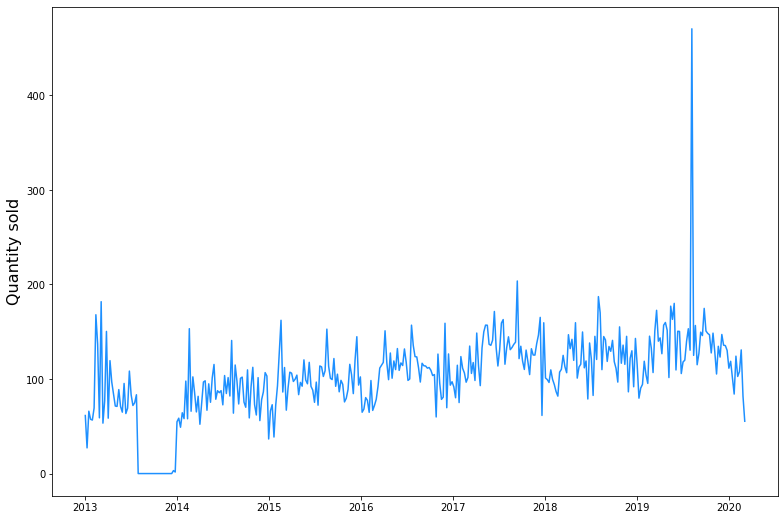}
  \end{subfigure}
  \caption{\textbf{(a)} Unit price and \textbf{(b)} sold quantity time series for product 2 after resampling with weekly frequency.\label{fig:p2_ts}}
\end{figure}

\begin{figure}[!h]
  \centering
  \begin{subfigure}[b]{0.48\textwidth}
    \centering
    \includegraphics[width=8.5 cm]{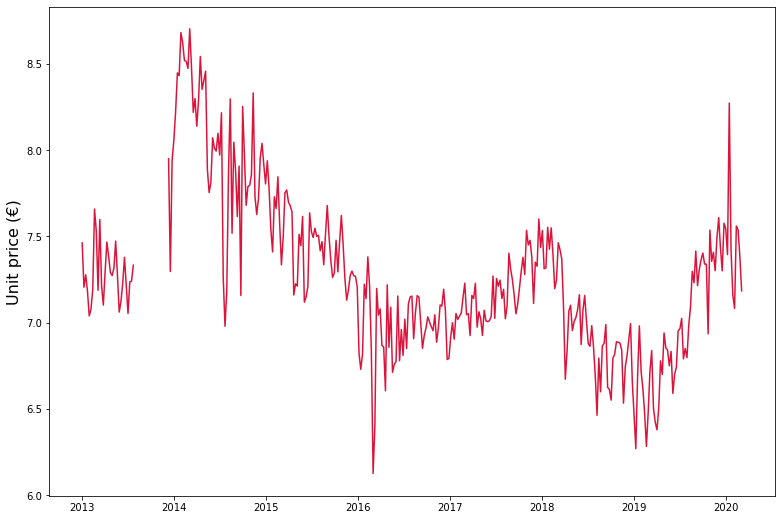}
  \end{subfigure}
  \hfill
  \begin{subfigure}[b]{0.48\textwidth}
    \centering
    \includegraphics[width=8.5 cm]{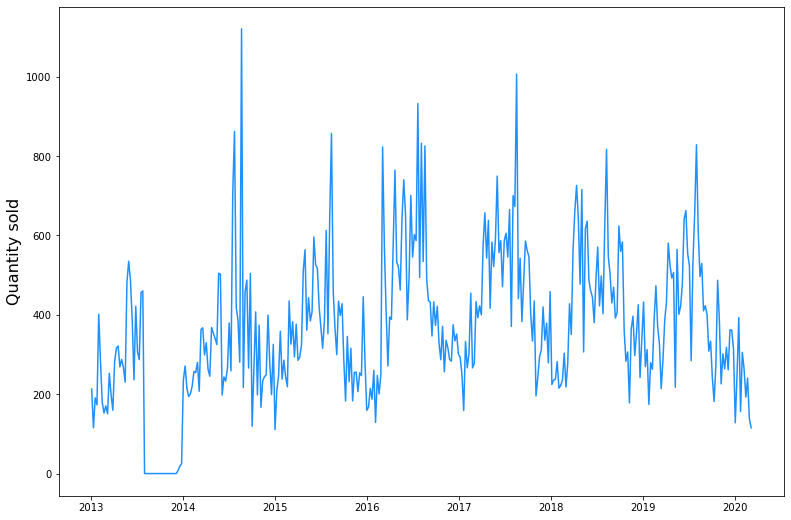}
  \end{subfigure}
  \caption{\textbf{(a)} Unit price and \textbf{(b)} sold quantity time series for product 3 after resampling with weekly frequency.\label{fig:p3_ts}}
\end{figure}

In view of the subsequent time series forecasting and as a further step in dealing with data inaccuracies, we decided to resample the dataset with weekly frequency.
This was done by cumulating the number of orders in each window and calculating the average sale price for each week. 
For later use in neural network models, when resampling data we also kept track of the following fields in the dataset: \emph{number served customers}, \emph{number of orders}, \emph{number of orders on sale}, (weighted) \emph{average product cost}, and (weighted) \emph{price standard deviation}. 
Table~\ref{tab:prices-stats} summarizes the main features of the resampled price time series for each of the products.
In Figures~\ref{fig:p1_ts}, \ref{fig:p2_ts} and \ref{fig:p3_ts} we display the time series of sale prices and sold quantities after resampling.
All prices, here and everywhere in the paper, are intended in euros (\euro).
\begin{table}[h]
  \caption{Mean and standard deviation of the weekly prices time series.}
  \label{tab:prices-stats}
  \centering
   \begin{tabular}{c c c c} 
   \toprule
   Product & 1 & 2 & 3 \\
   \midrule
   Mean (\euro) & 1.99 & 7.64 & 7.27  \\
   Std (\euro) & 0.36 & 0.40 & 0.26 \\
   \bottomrule
   \end{tabular}
\end{table}
We then split the weekly dataset in the following way:
\begin{itemize}
	\item \emph{Training dataset}: years 2013--2017,
	\item \emph{Validation dataset}: year 2018,
	\item \emph{Test dataset}: years 2019--March 2020
\end{itemize}
This choice was made in order to make the training set as large as possible, while also having validation and test sets that cover at least one year of data.

As can be seen in Figures~\ref{fig:p1_ts}, \ref{fig:p2_ts} and \ref{fig:p3_ts}, even after resampling the price time series have missing periods.
Therefore, they are not suited for being given as input to ARIMA and neural network forecasting models, as they require evenly spaced data.
To overcome this problem, we adopt the following strategy for all products:
\begin{enumerate}
  \item Since the time series for all products have a long window with no data in the second half of year 2013, we do not consider this empty period and start right after it;
  \item When occasional weeks with no data occur, we take the average of the preceding and following week prices and interpolate.
\end{enumerate}
In this way we were able to fill in all empty weeks -- in fact, after resampling the missing records were very sparse.
Note that the above procedure is only necessary for preparing the dataset for ARIMA and NN models, as Prophet has no problems in handling missing datapoints.
The size of the datasets for each product, both before and after removal of empty periods, is summarized in Table~\ref{tab:data}.
\begin{table}[h]
  \caption{Dataset size (\# of datapoints) for each product and forecasting model.}
  \label{tab:data}
  \centering
   \begin{tabular}{c  c  c  c  c  c} 
   \hline
    Product & Model & Train & Valid & Test  \\ 
   \hline
   \multirow{2}{*}{1}&  Prophet & 240 & \multirow{2}{*}{52} & \multirow{2}{*}{62}\\
   & ARIMA \& NN & 211 &  &   \\
   \hline
   \multirow{2}{*}{2}&  Prophet & 241 & \multirow{2}{*}{52} & \multirow{2}{*}{62}\\
   & ARIMA \& NN & 211 &  &   \\
   \hline
   \multirow{2}{*}{3}&  Prophet & 242 & \multirow{2}{*}{52} & \multirow{2}{*}{62}\\
   & ARIMA \& NN & 212 &  &   \\
   \hline
   \end{tabular}
\end{table}

\subsection{ARIMA models}
\label{sec:ARIMA-mod}

ARIMA models~\cite{box2015time} are among the most simple and used econometric approaches to univariate 
time series modeling.
In this work, we implemented non-seasonal ARIMA models, neglecting the modulation effects of holidays and using therefore pure trend lines.

In econometrics, it is quite customary when dealing with price variables to transform
prices through a logarithmic map, since this generally leads to better results. 
We decided to follow this approach when using the ARIMA modelling, thus working with $\log(p_t)$ in the model fitting.
As a first step we checked the stationarity properties of the time series. We performed the \emph{Augmented Dikey-Fuller}
unit root test using the built-in method in the \emph{statsmodels} Python package. 
The results we obtained are qualitatively similar for all the three products we considered: 
for the $\log(p_t)$  time series one cannot reject the null hypothesis of the presence of a unit root,
signalling the non-stationarity of the series. First differencing the series, \textit{i.e.} considering  
$\Delta \log(p_t) = \log(p_t) - \log(p_{t-1})$, makes it stationary. Thus the $\log(p_t)$ series 
are integrated of order one, and accordingly the models we considered are ARIMA$(p,1,q)$. 

\begin{figure}[h]
  \centering
    \includegraphics[width=.8\textwidth]{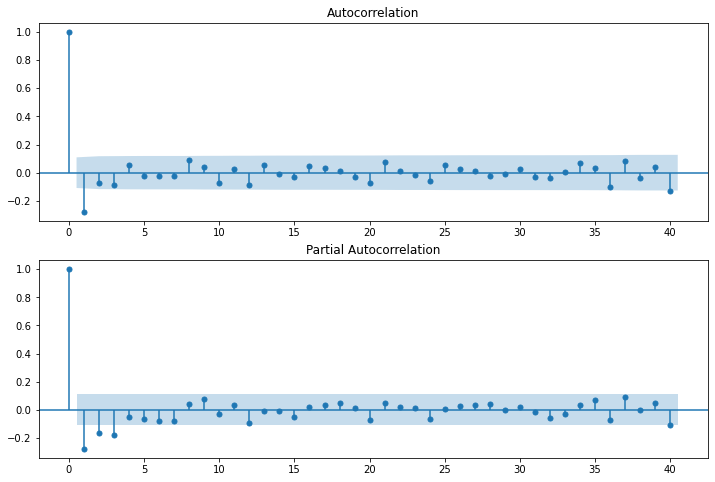}
   \caption{Autocorrelation function and partial autocorrelation function for $\Delta \log(p_t)$ 
   of product 2. }
   \label{fig:p1-ACF}
 \end{figure}

In order to have a rough indication on the AR orders, $p$'s, and on the MA orders, $q$'s, we computed 
the sample autocorrelation function (ACF) and the partial autocorrelation function (PACF)
for $\Delta \log(p_t)$. Recall that 
\begin{itemize}
	\item for an exact MA($q$), ACF is zero for lags larger than $q$;
	\item for an exact AR($p$), PACF is zero for lags larger than $p$.
\end{itemize}
As an example we show the plots of these functions for product 2 in Figure~\ref{fig:p1-ACF}.

In the ARIMA model selection and fitting procedures, we used a different dataset splitting scheme with respect to the one described above, in that the \emph{training} set comprised years 2013--2018 (i.e. the union of the former training and validation sets).
This is since we decided not to use the validation set to selected the hyperparameters $p$ and $q$, instead exploiting the \emph{Bayesian Information Criterion} (BIC)
as a metric for model comparison~\cite{james2013introduction}. Hence we took into account different combinations of 
$p$ and $q$ around the values suggested by the ACF and PACF plots, 
and eventually we selected the model with least BIC.

\subsection{Prophet}
\label{sec:proph-mod}
Prophet is an open-source tool provided by Facebook Inc., available both in Python and R. 
For the current analysis, the Python package (with Python 3.9) was used.
As explained by the authors~\cite{taylor2018forecasting}, the idea leading to Prophet was to develop a flexible forecasting tool which is easy to both use and tune.
The underlying model features a decomposable time series with three components: growth (or trend) $g(t)$, seasonality $s(t)$ and holidays $h(t)$ (if present).
In the present case, there are no obvious holidays to consider, as the wholesaler's customers are mainly restaurants and hotels, which tend to stay open during holidays.
The time series is therefore decomoposed as
\begin{equation}
	y(t)=g(t)+s(t)+\epsilon_t\, ,
\end{equation}
where $\epsilon_t$ encodes variations that are not taken into account by the model, and which are assumed to be normally distributed \cite{taylor2018forecasting}.
The Prophet model can be seen as a GAM \cite{hastie1987generalized}.
In this framework, forecasting is phrased as a curve-fitting task, with time as the only regressor, so the model is univariate.

The trend function adopted for the problem under study is a piecewise linear function written as
\begin{equation}
	g(t)=\big(k + \sum_{i: t > s_i}\delta_i \big) t + \big(m + \sum_{j: t > s_j} \gamma_j \big)\, ,
\end{equation}
where $k$ is a scalar coefficient, $s_i$ are the \emph{trend changepoints} -- \emph{i.e.} $S$ times $s_1$, $s_2$, ... , $s_S$ at which the angular coefficient of the trend is allowed to change -- $\delta_i $ are the rate adjustements, and $\gamma_j = -s_j \delta_j$ are parameters used to make the function continuous.
The algorithm starts with a number $S=25$ of potential changepoints, placed in the first $80\%$ of the time series in order to avoid responding to fluctuations in the last part of the series.
Then, the actual changepoints are selected by putting a sparse prior of the kind $\delta_j \sim \text{Laplace}(0, \tau)$, with $\tau$ (tunable hyperparameter) regulating\footnote{Using a Laplace prior is equivalent to L1-regularization.} the magnitudes rate adjustments. 
A larger $\tau$ means the model has more power to fit trend changes.

As for seasonality, Prophet accounts for it using Fourier series, namely
\begin{equation}
	s(t) = \sum_{n=1}^N \left( a_n \cos\left(\frac{2\pi n t}{P}\right)+ b_n \sin\left(\frac{2\pi n t}{P}\right)\right)\, .
\end{equation}
Since we considered weekly data with no other obvious expected seasonality effects but yearly ones, we had $P=365.25\text{d}$. 
For weekly seasonality, the truncation parameter is set to $N=10$ by the authors of \cite{taylor2018forecasting} when modelling yearly seasonality, and we follow this specification.
When performing the fit, a smoothing prior $\beta \sim N(0,\sigma^2)$ is imposed on the $2N$ components $\beta=(a_1,\ldots,a_n,b_1,\ldots,b_n)^T$, with $\sigma$ a second hyperparameter (essentially acting as an L2-regularization parameter).

Prophet fits its GAM using the L-BFGS quasi-Newton optimization method of \cite{byrd1995limited} in a Bayesian setting, finding a maximum \emph{a posteriori} estimate.

\subsection{Neural Networks}
\label{sec:NN-mod}

The advent of artificial intelligence, in particular machine learning, has led to the development of a set of techniques that have proved to be very useful in many different areas.
One breakthrough has certainly been deep learning~\cite{lecun2015deep}, which has revolutionized our way of handling and exploiting information contained in data.
Deep learning can effectively detect and model hidden complexity in data, automatically extracting features that should otherwise be extracted manually by dataset inspection. 

A standard choice when facing problems involving time series is that of using LSTM neural networks, a kind of recurrent neural networks (RNNs) devised by Hochreiter and Schmidhuber in 1997 \cite{hochreiter1997long}.
Like all RNNs, they can by construction handle data endowed with temporal structure, while also providing a way to deal with the vanishing gradient problem \cite{pascanu2013difficulty}.
Here we will describe the application of LSTM NNs to the problem under study, both on their own and in combination with CNNs.
Indeed, standard LSTM NNs for time series forecasting can be enriched with one-dimensional convolutional layers that sequentially apply a unidimensional filter to the time series.
Convolutions can be seen as non-linear transformations on the time series data.
This enhances the model's capability to learn discriminative features which are useful for the forecasting and that can be fed to the LSTM layers that follow.
The models developed in this work were trained and tested using Python 3.9 and TensorFlow 2.5.

Unlike in the ARIMA and Prophet case, with NNs we can exploit a larger fraction of the information available in the dataset by setting up a multivariate regression. 
However, since dates cannot be used as input variables to a NN, we made an addition to the fields listed in Section \ref{ssec:dataprep}, performing a time embedding to provide information about seasonality.
We did this by adding the columns
\begin{subequations}
	\begin{align}
		\text{week\textunderscore cos} &= \cos(2\pi w/52.1429)\, , \\
		\text{week\textunderscore sin} &= \sin(2\pi w/52.1429)\, ,
	\end{align}
\end{subequations}
where $w$ is the week number ($w=0,1,\ldots,52$).
Therefore, we had a total of 9 input columns that were passed to the NN models.
A sample of the input dataset for one product is shown in Table \ref{tab: dfNN}.
\begin{table}[h]
\caption{A slice of the dataset used to generate input data for the NN models.}
\label{tab: dfNN}
\centering
  \begin{tabular}{c c c c c c c c c c}
    \toprule
        quantity &   customers &   orders &   on sale &  cost &  week\textunderscore cos &  week\textunderscore sin & p\textunderscore std & avg\textunderscore price  \\
    \midrule
                10 &           1 &       1 &        0 &    1.40 &         0.990 &    -0.141 &    0 &  1.41  \\
                 0 &           0 &       0 &        0 &    1.40 &        1.000 &   -0.0214 &    0 &  1.55  \\
                70 &           6 &       6 &        0 &    1.40 &    0.993 &      0.120 &  0.0690 &    1.69  \\
               220 &          17 &      18 &        0 &    1.40 &   0.971 &     0.239 &   0.0580 &    1.75  \\
               230 &          14 &      15 &        0 &     1.39 &  0.935 &     0.353 &    0.0685 &  1.86  \\
    \bottomrule
    \end{tabular} 
\end{table}
To construct the actual training dataset (made of a tensor $\bf{x}_t$ and a scalar $y_t$, for each time $t$) that could be fed to LSTM neural networks, we then performed the following steps:
\begin{itemize}
  \item reshape data so that at each time $t$, $\bf{x}_t$ is a $n \times 9$ tensor containing the $n$ last values of each time series in Table \ref{tab: dfNN};
  \item set $y_t = \Delta_{t+1}=p_{t+1}-p_t$ as the variable to be used in the cost function.  
\end{itemize}
In this way, the model learns to predict $y$ (price variation) at each time based on information about the last $n$ timesteps.
Predicting the increment of the quantity of interest instead of the quantity itself is a well-known way to improve performance when training multivariate machine learning models.
Moreover, we checked through the \emph{Augmented Dikey-Fuller} test that the $\Delta_t$ time series was stationary.
The number $n$ of timesteps used depends on the model and will be specified later.

The NN models tried in this paper for predicting product prices fall in two classes: those using only LSTM layers and those with CNN layers before the LSTM.
We denoted these classes A and B, respectively.


\section{Results}
\label{sec:results}

In this section we report results obtained with the three different approaches to forecasting -- namely ARIMA, Prophet and deep learning -- studied in this work.

\subsection{ARIMA results}

We considered first ARIMA models, to provide a standard performance 
benchmark with which to compare the other models developed in the rest of this work.

The best ARIMA models for the three products are given in Table~\ref{tab:ARIMA-selected}.
As outlined in Section~\ref{sec:ARIMA-mod}, they were selected by considering the series of the price logarithms $\log(p_t)$ and using a least BIC criterion~\cite{james2013introduction}.
For the sake of comparison with the other models, we transformed back the $\log(p_t)$ series to the $p_t$ series to compute the root mean squared error (RMSE) between the predicted and observed increments
\begin{equation}
  \label{eq:Delta}
\hat{\Delta}(t)=\hat{p}_t - p_{t-1}\, , \qquad \Delta(t)=p_t - p_{t-1}\, ,
\end{equation}
$\hat{p}_t$ being the predicted price at time $t$.
\begin{table}[h] 
  \caption{Selected ARIMA models for the three products and associated RMSE on the entire training+validation set.}
  \label{tab:ARIMA-selected}
  \centering
  \begin{tabular}{c c c} 
    \toprule
    Product &  selected model & (tr+v)RMSE \\  
    \midrule
    1 & ARIMA(2,1,0) & 0.097\\
    2 & ARIMA(0,1,2) & 0.232\\
    3 & ARIMA(3,1,1) & 0.211\\
    \bottomrule
  \end{tabular}
\end{table}
In all cases, we checked also that the Ljung-Box statistics~\cite{box1970distribution,ljung1978measure} for 1-,6- and 12-lag residual autocorrelations do not reject the null hypothesis, 
so the residuals can be considered approximately white noise.

The results obtained on the test set by the selected ARIMA models are summarized in Table~\ref{tab:ARIMA-results}.
We also report values for the MAE (mean absolute error) and MAPE (mean absolute percent error).
\begin{table}[h] 
  \caption{Performance of the best ARIMA models on the test set. The MAPE is computed for price time series $p_t$, not for the $\Delta(t)$ time series. }
  \label{tab:ARIMA-results}
   \centering
   \begin{tabular}{c c c c } 
    \toprule
    \multicolumn{4}{c}{ARIMA results} \\
    \midrule
    Product  &  1 & 2 & 3  \\  
    \midrule
    RMSE & 0.0758 & 0.173 & 0.215 \\
    MAE & 0.0581 & 0.132 & 0.159\\
    MAPE & 0.0348 & 0.0178 & 0.0222 \\
    \bottomrule
  \end{tabular}
\end{table}

\subsection{Prophet results}

\subsubsection{Prophet grid search}

As suggested in the Prophet documentation and reviewed in Section~\ref{sec:proph-mod}, one can tune the $\tau$ (trend changepoints prior scale) and $\sigma$ (seasonality prior scale) hyperparameters so that the model fits data as well as possible.
We did so by performing a grid search over $\tau$ and $\sigma$ in the following way: for each combination of $\tau \in \{0.005, 0.01, 0.05, 0.1, 0.5\}$ and $\sigma \in \{0.01, 0.05, 0.1, 0.5, 1, 2\}$, we started by fitting the model over the training dataset, and predicted the price for the following week (first datapoint in the validation set).
We calculated the squared error between predicted and observed price. 
Then, we moved on to the second datapoint in the validation set, performed a new fit using also the first validation set datapoint, and predicted the price for the following week. 
The whole process was repeated until the validation dataset was exhausted.
For each product, the configuration with least RMSE was selected, yielding the results shown in Table \ref{tab:Proph-grid}.
\begin{table}[h]
  \caption{Prophet grid search results. vRMSE indicates the RMSE on the validation set, calculated as described in the main text.}
  \label{tab:Proph-grid}
    \centering
   \begin{tabular}{c c c c} 
   \toprule
   Product & $\tau$ & $\sigma$ & vRMSE  \\ 
   \midrule
   1 & 0.5 & 0.01 & 0.0831 \\ 
   2 & 0.1 & 0.01 & 0.293 \\ 
   3 & 0.5 & 1.0 &  0.215  \\ 
   \bottomrule
   \end{tabular}
  \end{table}

\subsubsection{Prophet forecasting}
\label{ssec:proph-res}

\begin{figure}[h]
	\centering
    \includegraphics[width=13 cm]{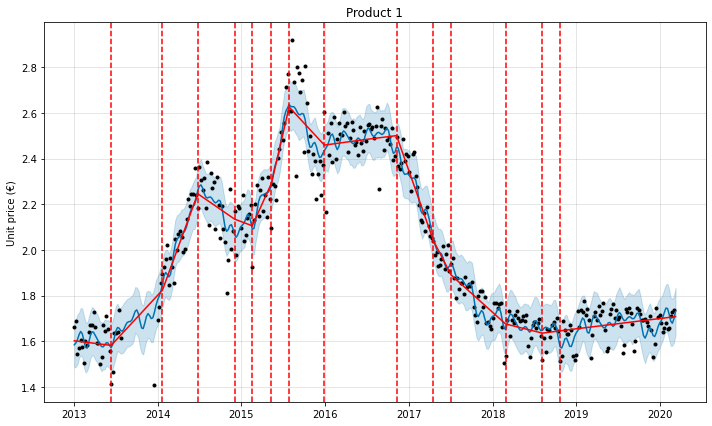}
    \caption{Prophet fit over the entire product 1 dataset. The actual fit is shown in light blue, while trend changepoints are shown in red.}
    \label{fig:p1-prophet}
  \end{figure}

\begin{figure}[h]
  	\centering
      \includegraphics[width=13 cm]{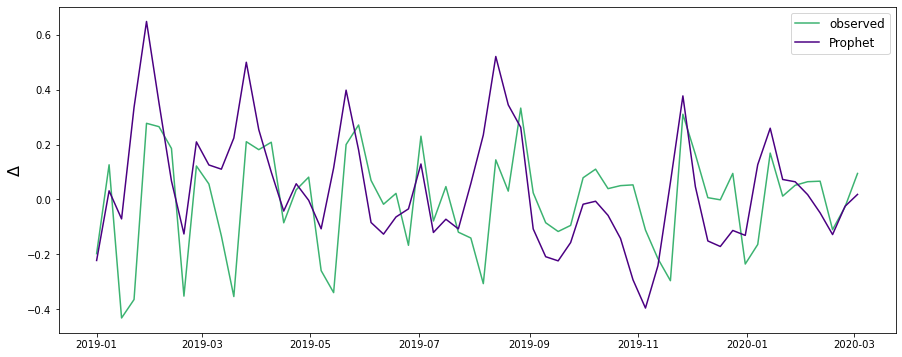}
    \caption{Comparison of Prophet forecasts and observed data for price variations in the test set, for product 2. The corresponding RMSE is 0.220.}
    \label{fig:p2-prophet-Delta}
\end{figure}

After selecting the best values of the parameters for each product, we employed them to specify the Prophet model in the test phase.
Testing took place in the following way: we started by fitting the model over the entire training plus validation dataset, predicting the first data entry in the test dataset and calculating the squared error.
We repeated the procedure for all entries in the test dataset, each time using all previous history, and calculated the achieved RMSE at the end of the process.
Note that by employing this procedure, also the test dataset is progressively used to fit the model.
We plot in Figure~\ref{fig:p1-prophet} the result of the fit over the entire dataset for product 1, i.e. the function that would be used to predict the unit price for the first week following the end of the test dataset.
Figure~\ref{fig:p2-prophet-Delta} shows instead the plot of the predicted and observed increments -- $\hat{\Delta}(t)$ and $\Delta(t)$ as defined in Eq.~\ref{eq:Delta} -- in the case of product 2.
The performance of Prophet in forecasting the weekly price time series is summarized in Table~\ref{tab:proph-results}.
As done for ARIMA models, besides the RMSE parameter used in the fine tuning, we also report values for the MAE and MAPE.
\begin{table}[h]
  \caption{Performance of the tuned Prophet models on the test dataset.}
  \label{tab:proph-results}
    \centering
    \begin{tabular}{c c c c} 
      \toprule
      \multicolumn{4}{c}{Prophet results} \\
      \midrule
    Product &  1 &  2 &  3 \\ 
      \midrule
      RMSE & 0.0812 & 0.220 & 0.350 \\ 
       MAE & 0.0694 & 0.165 & 0.301  \\
       MAPE & 0.0414  & 0.0224 & 0.0424 \\
      \bottomrule
    \end{tabular}
\end{table}
  
\subsection{Neural Networks results}

\subsubsection{NN grid search}
As a third forecasting tool, we studied deep neural networks.
Two different classes of models -- class A and B as introduced in Section~\ref{sec:NN-mod} -- were analyzed.
We performed two different grid searches to select the best model in each class, as we now briefly describe.
The common features between the classes were: the usage of an MSE cost function and of the \emph{Adam} optimization algorithm \cite{kingma2014adam}, with learning rate (or \emph{stepsize}) $\alpha \in \{0.0005, 0.001\}$; the adoption of an early stopping procedure, monitoring the cost function on the validation set with a patience of 5 epochs, while also setting an upper bound of 150 training epochs.
Finally, data were standardized using a \emph{z-score} normalization\footnote{Specifically, the \emph{MinMaxScaler} function of the \emph{scikit-learn} Python package was employed.}, and we used a batch size of 32.

\begin{figure}[h]
  \centering
    \includegraphics[width=11 cm]{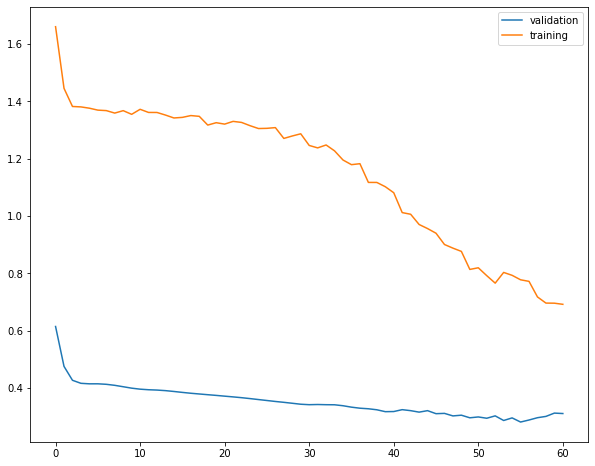}
    \caption{Trajectory of training and validation losses, as a function of the training epoch, for product 1. Note that the cost functions are calculated on rescaled data, therefore they cannot be directly compared to the values appearing on the first line of Table~\ref{tab:LSTM-CNN-grid}.}
    \label{fig:p1-trajectory}
\end{figure}

For class A, we trained NN models with the following architecture and hyperparameters: a number $l \in \{1,2,3\}$ of LSTM layers with $n_u \in \{ 32, 64, 96\}$ neurons each and \emph{normal Glorot} weight initialization \cite{glorot2010understanding}.  
Each LSTM layer was followed by a dropout layer, with dropout rate $r \in \{0.1,0.2,0.3\}$.
The output layer consisted of a single neuron with linear activation function, again with normal Glorot initialization.
For this class of models, we used a number of timesteps $n=4$.
The grid search over the hyperparameters $l$, $n_u$, $r$, $\alpha$ was performed by initializing and training each model ten times, monitoring the cost function on the validation set and recording the best result obtained for every configuration.
The best performing models for each product are reported in Table~\ref{tab:LSTM-grid}.
\begin{table}[h]
  \caption{Results of the grid search on class A (LSTM only). trRMSE and vRMSE are the RMSEs computed on the training and validation sets, respectively.}
  \label{tab:LSTM-grid}
    \centering
    \begin{tabular}{c c c c c c c} 
    \toprule
    Product & $l$ & $n_u$ & $r$ & $\alpha$ & trRMSE & vRMSE \\ 
    \midrule 
      1 & 3 & 32 & 0.1 & 0.001 &  0.0964 & 0.0612 \\ 
      2 & 3 & 64 & 0.3 & 0.001 & 0.163 &  0.173 \\ 
      3 & 1 & 96 & 0.1 & 0.0005 & 0.156 & 0.142 \\ 
    \bottomrule
    \end{tabular}
\end{table}
The second class of models, class B, consisted in a combination of CNN layers and LSTM layers, as done for example in~\cite{livieris2020cnn} and~\cite{hao2020predicting}.
We added two one-dimensional convolutional layers, with a pooling layer in between.
Each of the convolutional layers had $f \in \{10,20,30\}$ \emph{output filters}, \emph{kernel size} $k_s \in \{2,4\}$ and \emph{relu} (rectified linear unit) activation function.
Moreover, we tried to use \emph{same}, \emph{causal} or no padding in each of the conv1D layers: we dub the corresponding hyperparameters pad$_1$, pad$_2$.
The 1D \emph{average pooling} layer had \emph{pool size} equal to 2 and no padding. 
This block was then followed by the same LSTM layers as for the first class of models.
The reason behind adding CNN layers is that they can help make better use of the data history, improving the ability of LSTM networks to learn from long series of past data.
The grid search was indeed performed with varying numbers of timesteps $n \in {4,8,12}$.
As for the previous class, each model was initialized and trained twice.
Results for the grid search over the hyperparameters $l$, $n_u$, $r$, $\alpha, f, k_s, n$ are shown in Table~\ref{tab:LSTM-CNN-grid}.
Figure~\ref{fig:p1-trajectory} shows the trajectory of training and validation cost functions for the best-performing model in the case of product 1.
\begin{table}[h] 
  \caption{Results of the grid search on class B (CNN + LSTM). trRMSE and vRMSE are the RMSEs computed on the training and validation sets, respectively.}
  \label{tab:LSTM-CNN-grid}
  \centering
  \begin{tabular}{c c c c c c c c c c c c} 
   \toprule
  Product & $l$ & $n_u$ & $r$ & $\alpha$ & $f$ & $k_s$ & pad$_1$ & pad$_2$ & $n$ & trRMSE & vRMSE \\ 
   \midrule 
    1 & 1 &  64 & 0.3 & 0.0005 & 20 & 2 & causal & causal & 8 & 0.0770 & 0.0553 \\ 
    2 & 3 & 64 & 0.1 & 0.0005 & 20 & 2 & no & same & 12 & 0.175 & 0.165  \\ 
    3 & 1 &  32 & 0.2 & 0.001 & 20 & 2 & same & same & 8 & 0.148 & 0.132\\ 
    \bottomrule
   \end{tabular}
\end{table}

\subsection{NN forecasting}

\begin{figure}[h!]
  \centering
    \includegraphics[width=13 cm]{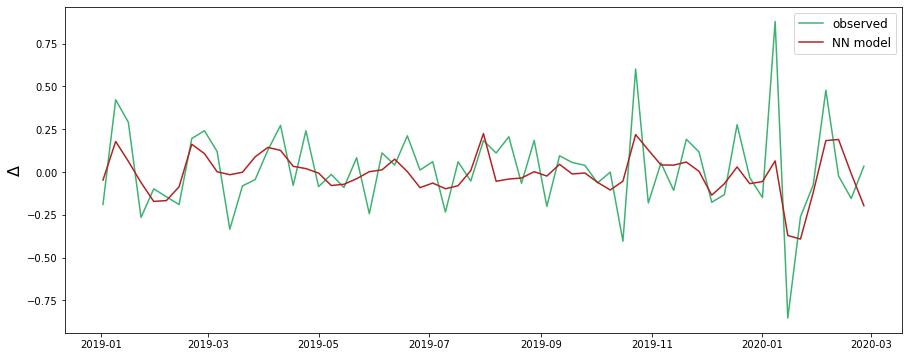}
  \caption{Comparison of NN forecasts and observed data for price variations in the test set, for product 3. The corresponding RMSE is 0.200.}
  \label{fig:p3-NN-Delta}
\end{figure}

We then used the models specified by the hyperparameter choices in Tables~\ref{tab:LSTM-grid} and~\ref{tab:LSTM-CNN-grid} to forecast price variations on the test set.
We proceeded by training the each selected model on the entire training \emph{and} validation set, for a number of epochs corresponding to the epoch at which the training had stopped during the grid search.
Table~\ref{tab:NN-results} reports the test set performances of NN models obtained in this way.
Although they played no role in model training and validation, here we also report the values of the MAE and MAPE metrics obtained on the test set.
We already observe that class B models always outperform class A models in forecasting prices for all the three products.
Figure~\ref{fig:p3-NN-Delta} displays the forecasts made by the best NN model and compares it to the actual price variations, in the case of product 3.
\begin{table}[h]
  \caption{Performance of the fine-tuned NN models on the test dataset.}
  \label{tab:NN-results}
  \centering
  \begin{tabular}{c c c c c c c} 
    \toprule
    \multicolumn{7}{c}{NN results} \\
    \midrule
    Product & \multicolumn{2}{c}{1} & \multicolumn{2}{c}{2} & \multicolumn{2}{c}{3} \\ 
    \midrule
    class & A & B  &  A & B  & A &  B   \\ 
    RMSE &  0.0617 & 0.0613  & 0.179 & 0.162 & 0.219 & 0.200  \\
     MAE  &  0.0498      &   0.0511      &  0.135     &  0.126     &   0.157    &      0.150   \\
     MAPE &  0.0299      &    0.0305    &    0.0181    &  0.0168    &   0.0221    &      0.0212   \\
    \bottomrule
    \end{tabular}
\end{table}

\subsection{Result comparison}

We can now compare the results obtained with the various models that have been tried.
The comparison can be made by looking at Tables~\ref{tab:ARIMA-results}, \ref{tab:proph-results} and \ref{tab:NN-results}

We begin by noting that Prophet performances are considerably poorer than the ARIMA benchmark and neural networks.
On one hand, this could be expected as Prophet is based on a model with relatively low complexity that prioritizes ease of use and of tuning.
On the other, as mentioned in Section~\ref{ssec:proph-res}, Prophet was progressively fitted also on the test set, effectively using more data than the other models for the fit, so its performances have arguably benefited from that.
Nevertheless, one point that needs to be stressed is that we were able to provide Prophet with the entire weekly price time series comprising all of 2013 data, with no need for particular data pre-processing.
Therefore, although Prophet's performances were not brilliant for the problem studied in this paper, it could still be useful in certain contexts where quick, preliminary forecasts are needed.

Turning to the other models, we observe that those featuring both CNN and LSTM layers (class B) yielded the most accurate forecasts for all the three products.
At the same time, tuning them required by far the longest time (approximately 20 hrs for each product using a NVIDIA RTX 3060 graphic card) with moderate improvement on purely LSTM (class A) models: considering the RMSE metric, the improvement amounts to about 9\% for both product 2 and 3, while for product 1 the two classes yield similar results (class B has lower RMSE but higher MAE and MAPE).
On the other hand, class A models required a considerably lower computational effort (of the order of 30 minutes to select the model hyperparameters for each product).
Notice that once the grid search was concluded, training the best model on the entire training plus validation dataset required only a few minutes, both for class A and B models.

ARIMA models performed well if we take into account both the achieved values of the metrics and time necessary for tuning.
They were less accurate than class B models, yielding RMSEs that were 23\% higher RMSE for product 1, 7\% higher for product 2, and 8\% higher for product 3\footnote{Similar considerations apply if we instead look at MAE and MAPE metrics.}.
However, they required about the same tuning time as class A models, and performed better for product 2 and 3: the RMSE obtained by ARIMA models was 23\% higher for product 1, but 2\% lower for products 2 and 3. 
We remark that ARIMA is univariate, while a multivariate input dataset was used to train and test deep learning models: this highlights the effectiveness of the ARIMA approach for the problem under study, while at the same time suggesting that the additional data fields used to perform the multivariate analysis were not so informative for price prediction.
We add that machine learning models were tuned over a larger parameter space than the others: the search grids were made of 54 and 8978\footnote{Factoring out the three different choices of the timesteps number $n$, which are not really part of the NN models but refer to the way data is prepared, we are left with 2916 configurations for class B models.} hyperparameter configurations for class A and class B NN models respectively, versus a maximum of 5 ARIMA configurations and the 25 of Prophet.

Another important aspect to consider in comparing the models is that dataset size strongly affects the performance of machine learning models. 
To this regard, we note that the sizes of the datasets were not very large (just over 200 for each product, as seen in Table~\ref{tab:data}), hence one could expect especially the NN models performance to further improve when more historical data is made available.

\section{Discussion}
\label{sec:discussion}
In this paper, we have discussed the application of different methods to the forecast of wholesale prices.
We put a standard econometric model (ARIMA) side by side with two different approaches to time series forecasting.
These were rather diverse both in kind and complexity, going from a simple additive model using a piecewise linear trend and Fourier series seasonality (Prophet) to deep learning models featuring both convolutional and LSTM layers.
The findings showed that while Prophet was quick to set-up and tune, requiring no data pre-processing, it was not able to come close to the performance of the other, well-established time-series forecasting models.
Instead, we found that the best deep learning models performed better than ARIMA, but also required much longer times for the hyperparameter tuning. 

The work done in this paper can be extended in many directions.
First, it would be interesting to carry out a similar analysis also for the forecasting of sales, and to consider data with daily frequency instead of weekly.
Sales forecasting with higher frequency can indeed be relevant for wholesalers and retailers.
Second, a more refined version of the study would distinguish between different customers, as the selling strategies adopted by the wholesaler do certainly differ when dealing with customers of various kinds and sizes.
Customer profiling is an extremely interesting and active avenue of applied research, which can clearly enhance companies' returns.
Therefore, in the near future we plan to carry on the analysis by combining customer classification algorithms with time-series forecasting: understanding how and when each customer buys, how specific products are treated differently by different customers (identifying \emph{complementary} and \emph{substitute} goods for each of them), as well as relating the price elasticity of demand for different product/customers, are all aspects that could lead to economic benefits for the wholesaler.
To this end, one would want to compare further machine learning models to dynamic panel data where prices for each product and customer and price demand elasticity are considered.
The approach could lead to an essential gain in the accuracy of the forecast, and would be an important contribution to sales prediction analysis, which is becoming an increasingly important part of modern business intelligence~\cite{mentzer2004sales,zhang2004neural}.
Another aspect of sure relevance would be to evaluate algorithms and models on data including the effects of the COVID-19 outbreak, to both learn how the market has transformed and help modifying selling strategies in adapting to the current rapidly changing situation. 

The limits encountered in the application of the forecasting tools examined in this work encourage the evaluation of further models that could bring together the advantages of each approach.
Finally, it would be relevant to apply one of the most exciting recent advances in machine learning, namely the attention mechanism~\cite{vaswani2017attention}, to the forecasting task we have considered. 
Work on this particular topic is definitely attractive and recently made its appearance in this field~\cite{ekambaram2020attention}.


%


\vspace{20pt}

\paragraph{Note:}
The paper is an output of the project ``Cancelloni Big Data Analytics (BDA)'' at Cancelloni Food Service S.p.A.
Data used in this work are property of Cancelloni Food Service S.p.A. and cannot be disclosed.


\printbibliography

\end{document}